\documentclass[11pt,a4paper]{article}
\usepackage{times,latexsym}
\usepackage{url}
\usepackage[T1]{fontenc}

\usepackage{times}
\usepackage{latexsym}
\usepackage{booktabs}
\usepackage{multirow}
\usepackage{amssymb}
\usepackage{microtype}
\usepackage{footmisc}

\usepackage{graphicx}
\usepackage{colortbl}
\usepackage{tikz}
\usepackage{forest}
\usetikzlibrary{trees,positioning,shapes,shadows,arrows.meta}

\usepackage[draft,textsize=footnotesize,textwidth=15mm]{todonotes}

\definecolor{indigo}{HTML}{4B0082}
\definecolor{amber}{HTML}{cc4e00}
\definecolor{teal}{HTML}{008080}
\definecolor{grey}{HTML}{979ea8}

\usepackage[acceptedWithA]{tacl2021v1}

\usepackage{xspace,mfirstuc,tabulary}

\newif\iftaclinstructions
\taclinstructionsfalse 
\iftaclinstructions
\renewcommand{\confidential}{}
\renewcommand{\anonsubtext}{(No author info supplied here, for consistency with
TACL-submission anonymization requirements)}
\newcommand{\instr}
\fi

\iftaclpubformat 

\else

\fi

\title{A Survey of AI-generated Text Forensic Systems: \\ Detection, Attribution, and Characterization}

\author{
Tharindu Kumarage\textsuperscript{1} ~ 
Garima Agrawal\textsuperscript{1}\thanks{\ \ Equal Contribution.}  ~ 
Paras Sheth\textsuperscript{1$\ast$} ~ 
Raha Moraffah\textsuperscript{1} \\ 
\textbf{Aman Chadha}\textsuperscript{2,3}\thanks{ \ \ Work does not relate to position at Amazon.} ~  
\textbf{Joshua Garland}\textsuperscript{1} ~ 
\textbf{Huan Liu}\textsuperscript{1}\\
\textsuperscript{1}Arizona State University, USA \\
\textsuperscript{2}Stanford University, USA \\
\textsuperscript{3}Amazon GenAI, USA \\
\texttt{\{kskumara,gsindal,psheth5,rmoraffa,jtgarlan,huanliu\}@asu.edu} \\
\texttt{hi@aman.ai} 
}

\date{}

\begin{document}
\maketitle
\begin{abstract}

We have witnessed lately a rapid proliferation of advanced Large Language Models (LLMs) capable of generating high-quality text. While these LLMs have revolutionized text generation across various domains, they also pose significant risks to the information ecosystem, such as the potential for generating convincing propaganda, misinformation, and disinformation at scale. This paper offers a review of AI-generated text forensic systems, an emerging field addressing the challenges of LLM misuses. We present an overview of the existing efforts in AI-generated text forensics by introducing a detailed taxonomy, focusing on three primary pillars: detection, attribution, and characterization. These pillars enable a practical understanding of AI-generated text, from identifying AI-generated content (detection), determining the specific AI model involved (attribution), and grouping the underlying intents of the text (characterization). Furthermore, we explore available resources for AI-generated text forensics research and discuss the evolving challenges and future directions of forensic systems in an AI era. 
   
\end{abstract}

\section{Introduction}


The advent of Large Language Models (LLMs) like GPT-4~\cite{openai2023gpt4}, Gemini~\cite{team2023gemini}, and open-source variants such as Falcon~\cite{almazrouei2023falcon} and Llama 1\&2~\cite{touvron2023llama}, has significantly enhanced natural language generation capabilities. These advancements have made it possible to produce text that is not only grammatically correct but also highly persuasive, closely mirroring human-written content. The utility of these models spans various domains across journalism, academia, and social media, where they serve as powerful tools for streamlining content creation processes. However,  these models introduce substantial challenges, particularly in the realm of information integrity. There is a growing concern over the potential misuse of LLMs for generating and spreading misinformation, propaganda, and disinformation, thus undermining public trust and the foundations of democracy~\cite{doi:10.1126/sciadv.adh1850, goldstein2024persuasive}. 

\begin{figure}
    \centering
    \includegraphics[width=1\linewidth]{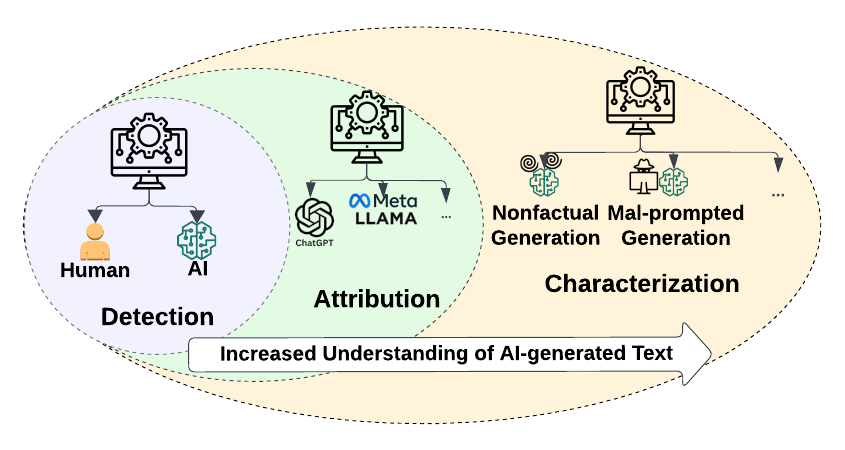}
    \caption{Primary pillars of AI-generated text forensics: (i) \textit{detection}, (ii) \textit{attribution}, and (iii) \textit{characterization}, where each pillar provides an increasingly nuanced understanding of AI-generated text.}
    \label{fig:intro}
    \vspace{-2.5mm}
\end{figure}

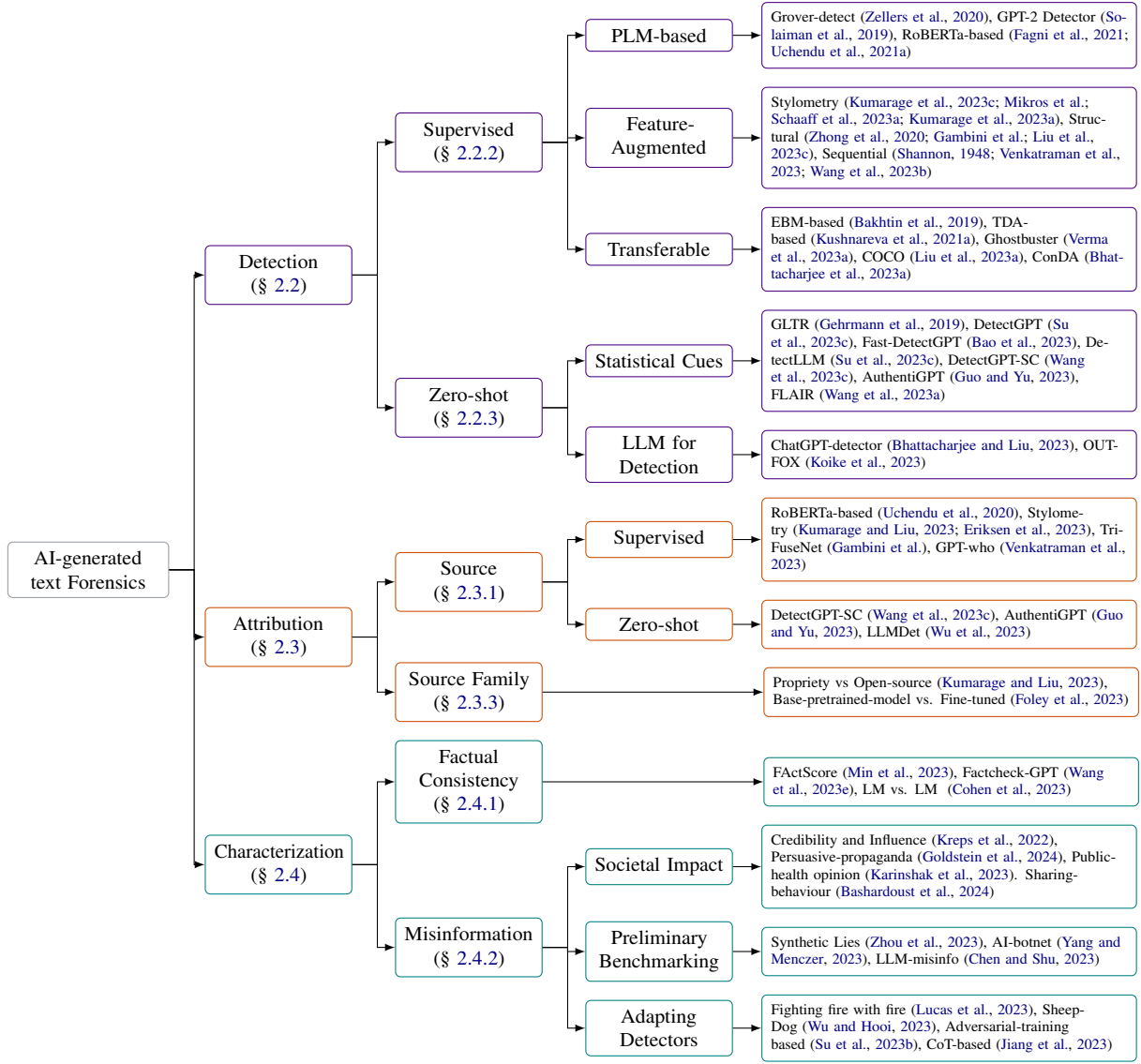
\begin{figure*}[t]
    \centering
\tikzset{
    basic/.style  = {draw, text width=2cm, align=center, rectangle, font=\scriptsize},
    root/.style   = {basic, draw=grey, rounded corners=2pt, thin, align=center, fill=none},
    infnode/.style = {basic, draw=indigo, thin, rounded corners=2pt, align=left, fill=none, text width=1.8cm, align=center},
    tranode/.style = {basic,draw=amber, thin, rounded corners=2pt, align=center, fill=none,text width=1.8cm},
    valnode/.style = {basic, draw=teal, thin, rounded corners=2pt, align=center, fill=none, text width=1.8cm},
    infcitenode/.style = {infnode, thin, align=left, fill=none, text width=50mm, font=\tiny},
    tracitenode/.style = {tranode, thin, align=left, fill=none, text width=50mm, font=\tiny},
    pretracitenode/.style = {tranode, thin, align=left, fill=none, text width=36mm, font=\tiny},
    valcitenode/.style = {valnode, thin, align=left, fill=none, text width=50mm, font=\tiny},
    edge from parent/.style={draw=black, edge from parent fork right}

}

\begin{forest} for tree={
    grow=east,
    growth parent anchor=west,
    parent anchor=east,
    child anchor=west,
    anchor=center,
    edge path={\noexpand\path[\forestoption{edge},->, >={latex}] 
         (!u.parent anchor) -- +(10pt,0pt) |-  (.child anchor) 
         \forestoption{edge label};}
}
[AI-generated text Forensics, root,  l sep=5mm,
    [Characterization (\S~\ref{sec:cha}), valnode,  l sep=6mm,
        [Misinformation (\S~\ref{sec:cha:misinfo}), valnode,l sep=6mm,
            [Adapting Detectors, valnode,l sep=4mm,
                [{Fighting fire with fire~\cite{lucas2023fighting}, SheepDog~\cite{wu2023fake}, Adversarial-training based~\cite{su2023fake}, CoT-based~\cite{jiang2023disinformation}}, valcitenode]
            ] 
            [Preliminary Benchmarking, valnode,l sep=4mm,
                [{Synthetic Lies~\cite{zhou2023synthetic}, AI-botnet~\cite{yang2023anatomy}, LLM-misinfo~\cite{chen2023can}}, valcitenode]
            ] 
            [Societal Impact, valnode,l sep=4mm,
                [{Credibility and Influence~\cite{kreps2022all}, Persuasive-propaganda~\cite{goldstein2024persuasive}, Public-health opinion~\cite{karinshak2023working}. Sharing-behaviour~\cite{bashardoust2024comparing}}, valcitenode]
            ] 
        ] 
        [Factual Consistency (\S~\ref{sec:cha:fact}), valnode,l sep=31mm,
            [{FActScore~\cite{min_factscore_2023}, Factcheck-GPT~\cite{wang_factcheck-gpt_2023}, LM vs. LM ~\cite{cohen_lm_2023}}, valcitenode]
        ] 
    ]
    [Attribution (\S~\ref{sec:att}), tranode,  l sep=6mm,
        [Source Family (\S~\ref{sec:att:fam}), tranode,l sep=31mm,
            [{Propriety vs Open-source~\cite{kumarage_neural_2023}, Base-pretrained-model vs. Fine-tuned~\cite{foley-etal-2023-matching} }, tracitenode]
        ] 
        [Source (\S~\ref{sec:att:clss}), tranode,l sep=6mm,
             [Zero-shot, tranode,l sep=4mm,
                [{DetectGPT-SC~\cite{wang_detectgpt-sc_2023}, AuthentiGPT~\cite{guo_authentigpt_2023}, LLMDet~\cite{wu2023llmdet}}, tracitenode]
            ] 
            [Supervised, tranode,l sep=4mm,
                [{RoBERTa-based~\cite{uchendu-etal-2020-authorship}, Stylometry~\cite{kumarage_neural_2023, eriksen_detecting_2023}, TriFuseNet~\cite{gambini_detecting_nodate}, GPT-who~\cite{venkatraman_gpt-who_2023}}, tracitenode]
            ]     
        ] 
    ]
    [Detection (\S~\ref{sec:det}), infnode,  l sep=6mm,
        [Zero-shot (\S~\ref{sec:det:zero}), infnode,l sep=6mm,
            [LLM for Detection, infnode,l sep=4mm,
                [{ChatGPT-detector~\cite{bhattacharjee_fighting_2023}, OUTFOX~\cite{koike_outfox_2023}}, infcitenode]
            ]
            [Statistical Cues, infnode,l sep=4mm,
                [{GLTR~\cite{gehrmann_gltr_2019}, DetectGPT~\cite{su_detectllm_2023}, Fast-DetectGPT~\cite{bao_fast-detectgpt_2023}, DetectLLM~\cite{su_detectllm_2023}, DetectGPT-SC~\cite{wang_detectgpt-sc_2023}, AuthentiGPT~\cite{guo_authentigpt_2023}, FLAIR~\cite{wang2023bot}}, infcitenode]
            ]
        ]
        [Supervised (\S~\ref{sec:det:sup}), infnode,l sep=6mm,
            [Transferable, infnode,l sep=4mm,
                 [{EBM-based~\cite{bakhtin2019real}, TDA-based~\cite{kushnareva-etal-2021-artificial}, Ghostbuster~\cite{verma_ghostbuster_2023}, COCO~\cite{liu_coco_2023}, ConDA~\cite{bhattacharjee_conda_2023}}, infcitenode]
            ]
            [Feature-Augmented, infnode,l sep=4mm,
                 [{Stylometry~\cite{kumarage2023stylometric, mikros_ai-writing_nodate, schaaff_classification_2023, kumarage_j-guard_2023}, Structural~\cite{zhong2020neural,gambini_detecting_nodate, liu_check_2023}, Sequential~\cite{shannon1948mathematical, venkatraman_gpt-who_2023, wang_seqxgpt_2023} }, infcitenode]
            ]
            [PLM-based, infnode,l sep=4mm,
                 [{Grover-detect~\cite{zellers2020defending}, GPT-2 Detector~\cite{solaiman2019release}, RoBERTa-based~\cite{fagni2021tweepfake, uchendu_turingbench_2021}}, infcitenode]
            ]
        ]
    ] 
]
\end{forest}
\vspace{-0.5mm}
\caption{Taxonomy of AI-generated text Forensic Systems.}
\vspace{-2.5mm}
\label{fig_tax}
\end{figure*}


Addressing these concerns necessitates a focused study of `AI-generated text forensics,' an emerging field dedicated to developing methodologies for analyzing, understanding, and mitigating the misuse of AI-generated text. This survey introduces the pillars of AI-generated text forensics as in Figure~\ref{fig:intro}: (i) \textit{detection}, (ii) \textit{attribution}, and (iii) \textit{characterization}—each serving a unique purpose in combating AI-generated content misuse. \textit{Detection} is pivotal for distinguishing between human and AI-generated texts, a fundamental step in safeguarding information integrity. \textit{Attribution} goes a step further by tracing AI-generated content back to its source model, thus promoting transparency and accountability. \textit{Characterization} seeks to understand the intent behind AI-generated texts, crucial for preempting harmful content.


To our knowledge, this is the first systematic review on AI-generated text forensic systems, featuring a detailed taxonomy as illustrated in Figure~\ref{fig_tax}. The necessity of this work stems from the evolving sophistication of AI-generated text and its potential misuse, requiring a multi-faceted approach for analysis and mitigation. Therefore, this survey aims to organize the current work, identify gaps and future directions in this rapidly developing field. Our work facilitates the advancement of research in AI-generated text forensics, contributing to the development of more robust, transparent, and accountable digital information ecosystems.\\
\noindent \textbf{Related Surveys:} Numerous surveys discuss aspects of detection~\cite{jawahar2020automatic, crothers2023machine, tang2023science} and attribution~\cite{uchendu2023attribution} in isolated contexts. In contrast, the objective of our survey is to delineate the broad themes within the AI-generated forensics field by identifying its fundamental pillars, exploring their interconnections, and discussing challenges envisioning a future where AI-generated text becomes pervasive.

\section{AI-generated Text Forensic Systems}
\label{sec:sys}

\subsection{AI-Generated Text}
\label{sec:pre}

In this survey, we define AI-generated text as output produced by a natural language generation pipeline employing a neural probabilistic language model~\cite{bengio2000neural}. 
The introduction of the transformer architecture~\cite{vaswani2017attention} was a critical milestone in the evolution of neural probabilistic language models, significantly enhancing sequential data processing. Transformers facilitate parallel processing and adeptly capture long-range dependencies in text. Consequently, these transformer-based LMs revolutionized the natural language generation process, enabling autoregressively querying it to generate the next token, given preceding context tokens. This breakthrough, coupled with advanced training techniques like instruction tuning and Reinforcement Learning from Human Feedback (RLHF) \cite{ouyang2022training}, laid the foundation for the creation of contemporary LLMs with extraordinary capabilities in generating grammatically correct, highly engaging text according to a given input prompt. 

\subsection{Detection Systems}
\label{sec:det}

In the field of AI-generated text forensics, detection models aim to determine if a text is authored by humans or generated by AI. This task is typically approached as a classification problem, wherein for a given text input $X$, the goal is to learn a function $d_\theta$ such that, $d_\theta(X) \rightarrow \{1,0\}$; where label $1$ indicates that the input text is AI-generated and $0$ implies human authorship.

\subsubsection{Watermarking vs. Post-hoc Detection}
Recent years have seen a surge in interest in developing AI text detection techniques, leading to a broad array of approaches that fall into two main categories: watermark-based and post-hoc detection. Watermarking involves embedding a detectable pattern into AI-generated text during training or decoding to later identify the text as originating from a specific LLM~\cite{ren_robust_2023, liu_survey_2024}. While effective, the application of watermarking is limited by the requirement of cooperation from the organization or the developer creating or hosting the LLM, a constraint not always met, especially with maliciously deployed LLMs. Consequently, post-hoc detection methods have gained prominence in AI-generated text forensics. Therefore, in the scope of our survey, we focus on post-hoc detection, further dividing it into supervised and zero-shot detection based on the training methodology employed.

\subsubsection{Supervised Detectors}
\label{sec:det:sup}
Supervised detectors are trained using annotated datasets that consist of labeled human-written and AI-generated texts, aiming to identify distinctive features between human and AI-generated writing.
\textls[-10]{Initial efforts in AI-generated text detection employed traditional techniques such as Bag-of-Words and TF-IDF encoding, coupled with classifiers like logistic regression, random forest, and SVC~\cite{ippolito2019automatic, jawahar2020automatic}. Subsequent research introduced advanced text sequence classifiers, including LSTM, GRU, and CNN, for detecting machine-generated text~\cite{fagni2021tweepfake}. A significant shift occurred when ~\cite{zellers2020defending} highlighted the impact of exposure bias in detecting text from LLMs, demonstrating that classifiers incorporating Grover layers achieved higher accuracy in identifying Grover-generated text. Therefore, subsequent advancements have focused on integrating pre-trained language models (PLMs) into classifiers, notably the OpenAI's GPT-2 detector~\cite{solaiman2019release, uchendu_turingbench_2021}, which utilizes a RoBERTa-based classifier trained on GPT-2 outputs~\cite{radford2019language}. Despite their effectiveness, these PLM-based detectors face challenges, including the rapid evolution of more sophisticated language models and the difficulty in transferring detectors across different models. To address these issues, recent approaches have explored feature augmentation to enhance classifier performance and the development of transferable methodologies that incorporate domain-invariant training strategies. The following sections detail feature-augmented and transferable approaches in supervised detection.}

\noindent \textbf{Stylometry Features}
Stylometry features serve as indicators of the nuances in writing styles between human and AI authors, based on the hypothesis that each exhibits distinct stylistic variations which can facilitate the detection of AI-generated text. Enhancing the pre-trained PLM-based classifiers with stylometric aspects such as phraseology, punctuation, linguistic diversity demonstrated improved performance in detecting AI-generated tweets~\cite{kumarage2023stylometric}. Subsequent research indicates that ensembles of stylometry features and PLM-based classifiers bolster the effectiveness of detection systems~\cite{mikros_ai-writing_nodate}. Beyond conventional stylometry attributes, ~\cite{schaaff_classification_2023} incorporates analysis of mean and maximum perplexity, sentiment, subjectivity, and error-based features like grammatical errors and the presence of blank spaces to enhance the detection capabilities. Journalism-standard features were introduced as a novel stylometry dimension, evaluating the compliance of news articles with the Associated Press Stylebook, to refine the accuracy of AI-generated news detection~\cite{kumarage_j-guard_2023}.

\noindent \textbf{Structural Features}
Various methodologies have been developed to enhance the capabilities of general detectors by incorporating explicit structural analysis of texts. ~\cite{zhong2020neural} improves detection accuracy by integrating the factual structure of text with a RoBERTa-based classifier~\cite{zhong2020neural}.  TriFuseNet~\cite{gambini_detecting_nodate}, a novel three-branched network was designed to explicitly model both stylistic and contextual features, thereby augmenting the detection of AI-generated tweets through fine-tuned BERTweet. Additionally,~\cite{liu_check_2023} improved detection capabilities by substituting traditional feed-forward layers with an attentive-BiLSTM in the classification head, enabling the classifier to discern between AI-generated and human-written texts through the learning of interpretable and robust features.

\noindent \textbf{Sequence-based Features}
Supervised methodologies investigate sentence-level or token-sequences to derive features grounded in information-theoretic principles~\cite{shannon1948mathematical}. For instance, GPT-who~\cite{venkatraman_gpt-who_2023} revisits the Uniform Information Density (UID) hypothesis, suggesting that unlike humans, who tend to distribute information uniformly during language production, AI-generated text may lack this evenness. Consequently, they introduce a set of UID features for quantifying the smoothness of token distribution. Similarly, SeqXGPT~\cite{wang_seqxgpt_2023} examines sentence-wise log probability metrics obtained from white-box LLMs to identify AI-generated text at sentence level. The authors draw an analogy of log probabilities to waveforms in speech processing and employ convolution and self-attention mechanisms to develop their classifier.

\noindent \textbf{Towards Transferable Supervised Detectors}

A well-recognized challenge with supervised detectors is their limited ability to generalize to novel AI generators. Various approaches have been explored to mitigate this issue, focusing on developing transferable techniques for AI-generated text detection. One such avenue involves integrating Energy-Based Models (EBMs) into the detection process~\cite{bakhtin2019real}. This integration exploits negative samples generated by multiple auto-regressive language models; specifically, the model assigns lower energy to human-generated text compared to text generated by AI models. Another strategy introduced by ~\cite{kushnareva-etal-2021-artificial} utilizes Topological Data Analysis (TDA) on attention maps produced by transformer models to extract domain-invariant features for AI-generated text detection. This approach involves representing attention maps as weighted bipartite graphs, leveraging TDA's capability to capture both surface and structural patterns in the underlying text.

\textls[-10]{More recently, \cite{verma_ghostbuster_2023} proposed Ghostbuster, a domain-generalized methodology employing three weak proxy language models to estimate token probabilities of the input text. This estimation is followed by a structured search over these token probability combinations. Subsequently, a linear classifier is trained on selected features to discern whether the input text is human-authored or AI-generated. Concurrently, the COCO framework~\cite{liu_coco_2023}, exploits inconsistencies in co-reference chains within AI-generated text as a domain-invariant feature. They enhance classifier representation by encoding entity consistency and sentence interaction within a supervised contrastive learning framework, with a focus on utilizing hard negative samples to boost model robustness. Additionally, ~\cite{bhattacharjee_conda_2023} introduced the ConDA model, which achieve transferability by incorporating standard domain adaptation techniques during training, by utilizing labeled training data from a source AI generator and unlabeled training samples from the target AI generator. ConDA integrates Maximum Mean Discrepancy (MMD) with the representational capabilities of contrastive learning to acquire domain-invariant representations, facilitating the adaptation of the classifier from the source generator to the target generator. Refer appendix Table \ref{tab:supervised_appendix} for detailed experiment settings.}

\subsubsection{Zero-shot Detectors}
\label{sec:det:zero}

Even though supervised detectors demonstrate state-of-the-art (SoTA) performance in the in-domain scenarios, they exhibit several shortcomings, such as a propensity to overfit the domain they were trained on and the necessity to train a new model for each newly released source AI generator. Given the rapid pace of current AI development, this becomes highly impractical. Consequently, recent extensive research has focused on devising zero-shot methods for AI text detection. Within the current literature on zero-shot detection, we identify two main categories: (1) detectors that leverage cues from LLM's probability function to differentiate human writing from AI writing and (2) those that employ LLMs directly as a zero-shot detector. 

\noindent \textbf{Cues from LLM’s Probability Function}
\quad \textls[0]{A notable characteristic of LLMs is their frequency bias; they are predisposed to select tokens prevalent in their training data when given a context. This contrasts with the diversity and surprise inherent in human writing~\cite{gehrmann_gltr_2019}. Motivated by this observation, several detectors were developed to leverage these probabilistic cues for zero-shot detection. 
GLTR~\cite{gehrmann_gltr_2019} employed a surrogate language model to assess the log probabilities of tokens within the text. It introduces statistical tests to determine the text's origin, whether AI or human, based on metrics such as average log probability, token rank, token log-rank, and predictive entropy.}

Subsequent research, such as DetectGPT~\cite{su_detectllm_2023}, empirically demonstrated that AI-generated text tends to be associated with negative curvature regions in the LLM's log probability function. Building on this insight, the authors proposed a text perturbation method to measure the log probabilities difference between original and perturbed texts. Here, a consistently positive difference suggests AI authorship. Fast-DetectGPT~\cite{bao_fast-detectgpt_2023} further streamlined this approach by eliminating the need for perturbation analysis and examining conditional probability curvatures, simplifying the detection process. This method revealed that AI-generated texts typically exhibit maximum conditional probability curvatures, unlike human-written text. Similarly, DetectLLM~\cite{su_detectllm_2023} found that AI texts have a higher Log-Likelihood Log-Rank Ratio (LRR) and are more affected by the Normalized Perturbed log-Rank (NPR) than texts written by humans.

Additional studies have explored the behavior of LLMs' probability function, focusing on the self-consistency aspect. The self-consistency posits that, given a specific input context, LLMs exhibit more predictable word or token selection in their responses compared to humans. Leveraging this concept, DetectGPT-SC~\cite{wang_detectgpt-sc_2023} introduced a detection method based on masked prediction. This technique involves masking certain words in the input text and asking the LLM to predict these words. A high degree of prediction consistency with the actual text suggests that the text was likely generated by the LLM in question. Similarly, AuthentiGPT~\cite{guo_authentigpt_2023} assesses the consistency aspect by applying a black-box LLM to denoise text that has been intentionally distorted with noise, then semantically comparing the denoised text against the original to ascertain if it is AI-generated. Another approach, proposed by~\cite{zhu_beat_2023}, is based on measuring the volume of text rewrites by ChatGPT. The underlying assumption is that the ChatGPT model requires fewer modifications to AI-generated texts than to those authored by humans.

\textls[-10]{Diverging from the above methodologies, FLAIR~\cite{wang2023bot} adopted an online bot detection strategy, which assumes query access to the AI generator in a black-box manner. Authors formulate a series of diagnostic questions and responses help distinguish whether the source is an AI or human by categorizing questions into those easily answered by humans but challenging for bots (e.g., counting, substitution, positioning, noise filtering, and ASCII art) and vice versa (e.g., memorization and computation).}

\noindent \textbf{LLMs as Zero-shot Detectors}
\textls[0]{Several studies have explored the potential of leveraging LLMs as zero-shot detectors in the field of AI-generated text detection. ~\cite{bhattacharjee_fighting_2023} conducted an analysis using GPT-3.5 and GPT-4 to automatically classify texts as either human-written or AI-generated. Their findings suggest that employing these models directly is not a reliable method for detection. OUTFOX~\cite{koike_outfox_2023} introduced a more effective strategy simulating an adversarial training environment through In-Context Learning. This approach involves a dual-system of a detector LLM and an attacker LLM. Initially, the detector LLM assigns labels to a training dataset. Subsequently, the attacker LLM crafts adversarial texts based on these initial labels. The detector LLM then utilizes these adversarially crafted texts as few-shot examples to enhance its ability to identify AI-generated content in a test dataset. Refer appendix Table \ref{tab:zeroshot_styled} for detailed experiment settings.}

\subsection{Attribution Systems}
\label{sec:att}

\textls[-10]{In the field of AI-generated text forensics, attributing the text to it's originating source LLM, termed neural authorship attribution, is crucial for augmenting the transparency of AI-generated text. This task is typically approached as a multi-class classification problem, wherein for a given text input $X$, the goal is to learn a function $a_\theta$ such that, $a_\theta(X) \rightarrow \{0,1, \dots, k-1\}$; where labels  $0,1, \dots, k-1$ indicates the $k$ known source generators.}

\subsubsection{History of Authorship Attribution}
\label{sec:att:clss}

Authorship attribution (AA), the task of recognizing authors by their unique writing styles, has been extensively studied for many years. Initially, classical classifiers such as Naive Bayes, SVM, Decision Trees, Random Forest, and KNN, along with feature extraction methods like n-grams, POS tags, topic modeling, and LIWC, were utilized to address AA challenges ~\cite{koppel2009computational, uchendu2023attribution}. Advancements in neural networks led to the adoption of Convolutional Neural Networks and Recurrent Neural Networks for AA, thanks to their capacity to capture an author's distinctive characteristics ~\cite{boumber-etal-2018-experiments, alsulami2017source}. The introduction of transformer-based models marked a significant evolution in AA, transitioning from traditional stylometric and statistical features to employing PLM-based classifiers~\cite{uchendu-etal-2020-authorship}. These classifiers have achieved SOTA performance in identifying neural authors as well.

\subsubsection{Extending Detection to Attribution}
\label{sec:att:det}
Both supervised detection and supervised attribution approaches share several common techniques. Stylometry-augmented PLM-based detectors, for example, have been directly applied to attribution tasks~\cite{kumarage_neural_2023, eriksen_detecting_2023}. Similarly, the TriFuseNet detection approach has proven effective in identifying source generators~\cite{gambini_detecting_nodate}. The information-theory-based GPT-who~\cite{venkatraman_gpt-who_2023} also demonstrates that the same UID features used in detection are relevant for neural AA.

Furthermore, many zero-shot detection methods discussed previously can be directly applied to neural AA tasks. Specifically, the detectors that incorporate self-consistency aspects, such as DetectGPT-SC~\cite{wang_detectgpt-sc_2023} and AuthentiGPT~\cite{guo_authentigpt_2023}, calculate consistency using a target LLM. For multiple source-LLMs, these approaches can assess consistency measures across sources to identify the most likely origin. Additionally, LLMDet~\cite{wu2023llmdet} proposed a perplexity-score comparison for neural AA. However, calculating perplexity requires white-box access to token-level log probabilities, which is impractical in real-world scenarios. Instead, they suggest calculating a proxy perplexity for each target LLM using common n-gram probabilities, serving as the LLM's writing signature to determine the closest source to the input text's proxy perplexity.

\subsubsection{Source Family Classification}
\label{sec:att:fam}

In addition to the general attribution methods previously discussed, there are techniques focused on tracing the attribution back to the base-model family. Such analysis is particularly valuable for inferring the budget and expertise behind malicious influence campaigns and determining which classes of LLMs are susceptible to these campaigns. ~\cite{kumarage_neural_2023} conducted a study to attribute LLM-generated text to high-level model families, such as 'proprietary' and 'open-source.' By integrating stylometry feature-augmented PLM-based classifier, they demonstrated that this task could be achieved with high accuracy. Further, ~\cite{foley-etal-2023-matching} shows how existing PLM-based attribution methods could identify the base-LLM from its fine-tuned variations, offering deeper insights into the origins of generated content.

\subsection{Characterization Systems}
\label{sec:cha}

\textls[-10]{Detection and attribution are crucial for identifying AI authorship, yet their primary limitation lies in their inability to provide insights into potential misuses or malicious intent behind the identified authorship. It is essential to ascertain whether AI-generated text harbors malicious intent to mitigate its harmful impacts effectively. We refer to the process of uncovering the intent behind AI-generated text as a fundamental aspect of AI-generated text forensics. At a high level, the task of characterizing intent can be conceptualized as a classification problem. Given a text input $X$, the objective is to learn a function $c_\theta$ that maps $X$ to $\{0,1\}$, with $0$ denoting non-malicious intent and $1$ indicating malicious intent. However, assessing intent from text is subjective and complex in practice, making this direct approach challenging~\cite{wang2023understanding, subbiah-etal-2023-towards}. Therefore, direct characterization of intent may currently seem ambitious. Yet, as we move towards an AI-centric future, characterization will become crucial in addressing AI-content misuse. Therefore, in this survey, we aim to review emerging directions foundational to characterization, including factual consistency evaluation and AI-misinformation detection, which will be discussed further in subsequent sections.}

\subsubsection{Factual Consistency Evaluation}
\label{sec:cha:fact}
Evaluating the factual accuracy of text produced by LLMs is a critical preliminary step in text characterization. Initially, human fact-checkers played a pivotal role in this process. However, the volume of text generated by contemporary LLMs has made manual verification methods increasingly impractical. This challenge has motivated the development of automated techniques for assessing the factual consistency of LLM-generated text.

\textls[-10]{For instance, FActScore~\cite{min_factscore_2023} introduces an innovative approach to evaluate the factual consistency of lengthy texts by deconstructing them into individual facts and verifying them against trustworthy sources. This method combines human judgment and automated processes, underscoring the efficiency and scalability of automated fact verification compared to traditional methods. Similarly, Factcheck-GPT~\cite{wang_factcheck-gpt_2023} provides an end-to-end system for verifying the facts of LLM outputs, employing a detailed annotation process and a customized tool to streamline the verification process. Additionally, ~\cite{cohen_lm_2023} presents a cross-examination framework that leverages interactions between different LMs to uncover factual discrepancies in LLM-generated texts.}

\subsubsection{AI-Misinformation Detection}
\label{sec:cha:misinfo}
\textls[-10]{The detection of misinformation generated by LLMs is crucial for characterizing and mitigating the misuse of AI-generated text. This area focuses explicitly on identifying AI-generated text that contains misinformation, differing from general AI-generated text detection by emphasizing the challenge of pinpointing deceptive information. Within AI-misinformation detection, several sub-categories of research have emerged in recent years. These include studies on the societal impact of AI-generated misinformation, which pose critical questions regarding its persuasiveness and dissemination compared to human-written misinformation. Additionally, there is a body of work focused on developing taxonomies for how adversaries might utilize LLMs to create misinformation and evaluating existing methods for detecting such content. Finally, some studies address adapting to the emerging threat of LLM-generated misinformation by proposing innovative detection mechanisms.}

\noindent \textbf{Societal Impact}\quad Early experiments evaluating the credibility and influence of AI-generated texts on foreign policy opinions have demonstrated that partisanship significantly affects the perceived credibility of the content. However, exposure to AI-generated texts appears to minimally impact policy views~\cite{kreps2022all}. This finding highlights the potential of AI to rapidly produce and disseminate large volumes of credible-seeming misinformation, thereby worsening the misinformation challenge in the news landscape, undermining media trust, and fostering political disengagement. Further research employing GPT-3 to generate persuasive propaganda has shown that such models can produce content nearly as compelling as that created by human propagandists~\cite{goldstein2024persuasive}. Through prompt engineering the effort required for propagandists to generate convincing content can be significantly reduced, underscoring AI's role in facilitating misinformation.

Moreover, researchers explore the impact of AI-generated texts on public health messaging, finding that AI-generated pro-vaccination messages were considered more effective and elicited more positive attitudes than those authored by human entities like the Centers for Disease Control and Prevention (CDC)~\cite{karinshak2023working}. A recent study delve into the sharing behavior and socio-economic factors affecting the spread of AI-generated fake news~\cite{bashardoust2024comparing}. They identify socio-economic factors, such as age and political orientation, as significant influencers of susceptibility to AI-generated misinformation. These findings suggest the necessity for customized media literacy education and regulatory measures to address the challenges posed by AI-generated misinformation.

\noindent \textbf{Preliminary Benchmarking}\quad The initial step in combating AI-generated misinformation involves examining how malicious actors exploit contemporary LLMs to produce such content. Consequently, numerous studies have recently developed taxonomies for AI-misinformation generation and established benchmarking datasets, in addition to evaluating the effectiveness of existing detectors against such LLM-generated misinformation.

Synthetic Lies~\cite{zhou2023synthetic} sets a benchmark for differentiating AI-generated misinformation from human-written news, focusing on a dataset related to COVID-19. This research uncovers linguistic patterns unique to AI-generated misinformation, such as enhanced detail and simulated personal anecdotes, challenging traditional detection models like CT-BERT. ~\cite{yang2023anatomy} analyzed a Twitter botnet that employs ChatGPT to disseminate misleading content. Their findings point out the limitations of current detection tools in identifying bot-generated text powered by LLMs. Furthermore, ~\cite{chen2023can} delve into the intricacies of detecting LLM-generated misinformation, offering a comprehensive taxonomy that includes generation methods (e.g., hallucination, arbitrary misinformation, and controlled misinformation generation), as well as the domains and intentions behind the misinformation. Their analysis reveals that misinformation crafted by LLMs poses more significant detection challenges, underscoring the urgent need for countermeasures.

\noindent \textbf{Adapting Detectors for AI-Misinformation}\quad Recent research has focused on enhancing detectors to address the challenges posed by AI-generated misinformation. For instance, Lucas et al.\cite{lucas2023fighting} propose a novel methodology that employs LLMs for both generating and detecting misinformation. Utilizing the generative prowess and zero-shot semantic reasoning capabilities of GPT-3.5-turbo, this approach significantly enhances the accuracy of distinguishing authentic content from deceptive information. Concurrently, SheepDog~\cite{wu2023fake},a style-agnostic detection system tackle the issue of LLMs being used to craft misinformation that mimics credible sources.

\textls[-10]{Moreover, ~\cite{su2023fake} highlights the inherent biases of existing detectors towards LLM-generated content and shows paraphrased example-based adversarial training as a mitigation strategy. A subsequent study reveals that while detectors trained on human-authored articles can somewhat identify machine-generated misinformation, the reverse is less effective \cite{su2023adapting}. This insight led to exploring how adjusting the ratio of AI-generated to human-written news in training datasets could enhance test-set detection accuracy. Additionally, Jiang et al.~\cite{jiang2023disinformation} offer an overview of the difficulties in identifying LLM-crafted disinformation, advocating for advanced prompting techniques, such as Chain of Thought (CoT) and contextual analysis, as viable strategies.}

\section{Resources}
\label{sec:res}

\begin{table*}[t]
\resizebox{\textwidth}{!}{%
\begin{tabular}{p{3cm}p{4cm}p{2.5cm}p{4.5cm}p{3.5cm}cp{2.75cm}} 
\toprule
\rowcolor[HTML]{CBCEFB} 
\multicolumn{1}{l}{\cellcolor[HTML]{CBCEFB}} & 
\multicolumn{6}{c}{\cellcolor[HTML]{CBCEFB}\textbf{Comparison Attributes}} \\ \cline{2-7} 
\rowcolor[HTML]{CBCEFB} 
\textbf{Model} & 
\textbf{Generators} &
\textbf{Domain} &
\textbf{Data Sources} &
\textbf{Training Samples} &
  \textbf{Metrics} & \textbf{\begin{tabular}[c]{@{}c@{}}Top  \\ Performance\end{tabular}} \\ 
\midrule
\rowcolor[HTML]{EFEFEF} 
\cellcolor[HTML]{EFEFEF} 
Facts from Fiction \cite{mosca2023distinguishing} & SCIgen, GPT-2, GPT-3, ChatGPT, Galactica & Scientific papers & arXiv & 16k - \textit{real}, 13k - \textit{fake}, 4k - \textit{para} & Acc & 77\%(OOD), 100\% (inDomain) \\
AuTexTification ~\cite{sarvazyan2023overview} & BLOOM (1B7, 3B, 7B1), GPT (babbage, curie, text-davinci003) & Tweets, News, Reviews, How-to articles, Legal & En (MultiEURLEX, Amazon Reviews, WikiLingua, XSUM, TSATC), Es (MLSUM, XLM-Tweets, COAR, COAH, TSD) & 160k texts & Macro-F1 & 80.91\%(En), 70.77\%(Es)\\
\rowcolor[HTML]{EFEFEF}
MULTITuDE \cite{macko2023multitude} & \textit{Multilingual LLMs}: GPT-3, GPT-4, LLaMA65B, ChatGPT, Vicuna-13B, OPT-66B, IML-Max-1.3B, Alpaca-LoRa-30B & News & MassiveSumm & 11 Languages: 74k (\textit{human written} - 8k, \textit{machine gen} - 66k) & Acc & 94.50\%\\
M4 \cite{wang2023m4} & ChatGPT, textdavinci-003, LLaMa, FlanT5, Cohere, Dolly-v2, BLOOMz & News, Scientific articles, Peer Reviews, Social Media, History, Web & En (Wiki, WikiHow, Reddit, arXiv), Chinese (PeerRead, Baike, WebQA), Urdu, Indonesian (News), Russian (RuATD) & 122k (En - 101k, \textit{other Languages} - 9k each) & F1 & 99.7\% \\
\rowcolor[HTML]{EFEFEF}
TURINGBENCH \cite{uchendu2021turingbench} & Transformer\_XL, PPLM, XLNET, Grover, CTRL, XLM, FAIR, GPT-1, GPT-2, GPT-3 & Politics, News & CNN, Washington Post & 10K - \textit{real}, 200k - \textit{machine gen} & F1 & 87.9\%(Detection), 81\%(Attribution)\\
\midrule
\multicolumn{7}{c}{\textbf{AI-Misinformation Benchmarks} $\downarrow$} \\
\midrule
LLMFake \cite{chen2023can} & ChatGPT, Llama2 (7b, 13b, 70b), Vicuna (7b, 13b, 33b) & News, Healthcare, Politics & \textbf{Pol}itifact, \textbf{Gos}sipcop, \textbf{CoA}ID & \textbf{Pol} (270-\textit{nonfactual}, 145-\textit{factual}), \textbf{Gos} (2230-\textit{nonfactual}), \textbf{CoA} (925-\textit{factual}) & Success Rate & Drops by 19\%\\
\rowcolor[HTML]{EFEFEF}
F3 \cite{lucas2023fighting} & GPT-3.5 & Political, News, Social Media & Politifact1, Snopes & \textit{human written}: (5508 - \textit{real}, 7215 - \textit{fake}), \textit{machine gen}: (9141 - \textit{real}, 18526 - \textit{fake}) & Acc & 72\% \\
ODQA-NQ-1500, CovidNews \cite{pan2023risk} & GPT-3.5 (text-davinci-003) & Web, News & Wiki Natural Questions, StreamingQA News & 21M (\textbf{NQ}), 3.3M (\textbf{Cov}) & Exact Match & 87\% Drop\\
\rowcolor[HTML]{EFEFEF}
Synthetic Lies \cite{zhou2023synthetic} & GPT-3.5 & News, Social Media (SM) & COVID19-FNIR, COVID Rumor, Constraint & 12k (6768 News, 5640 SM) & F1 & 98.5\%\\
\rowcolor[HTML]{EFEFEF}
\textbf{Gos}sipCop++, \textbf{Pol}itiFact++ \cite{su2023fake} & ChatGPT & News, Politics & FakeNewsNet, PolitiFact, GossipCop & 10k \textit{human written} (5k-\textit{real}, 5k-\textit{fake}, 5k- \textit{machine fake}) & Acc & 88\%\textbf{Gos++}, 80.93\% \textbf{Pol++}\\
\bottomrule
\end{tabular}%
}
\caption{Summary of Benchmark Datasets (\textit{En: English, Es: Spanish}).}
\label{tab:benchmark_datasets}
\end{table*}

\textls[-10]{Table \ref{tab:benchmark_datasets} offers an overview of set of significant datasets used in AI-generated text forensics, assessed across several crucial dimensions, including the AI generators used, the domains of writing, and performance metrics. These datasets fall into two main categories: general AI-generated text datasets (for detection and attribution purposes) and AI-misinformation datasets (for characterization).}

\subsection{Generators and Domains}
\textls[-10]{The datasets utilize a wide variety of generators, such as SCIgen, GPT models (GPT-2, GPT-3, GPT-3.5), BLOOM, and more, across a broad set of domains from scientific papers to social-media posts and academic works. For instance, Facts from Fiction~\cite{mosca2023distinguishing} focuses on scientific papers, drawing on sources like arXiv, whereas AuTexTification~~\cite{sarvazyan2023overview} covers domains such as tweets, reviews, and news articles. This diversity underscores the datasets' comprehensive coverage in testing detection and attribution systems.}

\subsection{Performance Metrics}
\textls[-10]{The benchmarks use metrics like accuracy and F1 scores to evaluate the effectiveness of detection and attribution. We highlight the top performance records for each dataset. Detection performance in general AI-generated text is notably high. For example, the MULTITuDE \cite{macko2023multitude} dataset, which concentrates on news text, marked an accuracy of 94\%. In contrast, AI-misinformation detection performance is significantly lower, reflecting the complex challenges inherent in characterizing AI-generated misinformation.}

\subsection{AI-Misinformation Benchmarks}
\textls[-10]{Specific benchmarks address the difficulty of detecting AI-generated misinformation. Notably, early benchmarks such as Synthetic Lies~\cite{zhou2023synthetic} demonstrate strong performance (95\%+), whereas more recent, complex, taxonomy-based benchmarks such as LLMFake~\cite{chen2023can} show weaker detection performance. This underscores the need for datasets that can mimic the sophisticated and evolving strategies of real-world misinformation campaigns. Through in-depth analysis of generation parameters, such as the use of particular mal-intent prompts, these datasets offer crucial insights for characterization systems.}

\subsection{Generation Parameters}
Additionally, the datasets shed light on generation parameters and multilingual support, tackling the worldwide challenge of AI-generated misinformation. Due to brevity, only key datasets are summarized in this table; a full list of benchmarks, complete with their specific generation parameters, seed prompts, and detailed performance metrics, can be found in Table \ref{tab:benchmark_appendix} in the Appendix.

\section{Future of AI-generated Text Forensics}
\label{sec:fut}
\textls[-10]{The rapid evolution of LLMs foreshadows an AI-centric future where AI systems may partially or entirely manage many everyday writing tasks. Concurrently, this shift introduces significant challenges and more complex threat scenarios. In the subsequent sections, we explore such potential challenges and envision future improvements for AI-generated text forensic systems.}

\subsection{Future Threat Landscape}

\subsubsection{Diminishing Boundary}
\textls[0]{A significant challenge is the blurring of distinctions between human-written and AI-generated text. Current detection systems operate on the premise that a discernible distribution shift exists between texts authored by humans and those produced by AI. However, recent advancements in LLMs have significantly improved their ability to mimic human writing styles. A theoretical analysis conducted in a recent study revealed that, for a sufficiently advanced language model aimed at imitating human text, the efficacy of even the most sophisticated detectors might only slightly surpass that of a random classifier~\cite{sadasivan_can_2023}. Consequently, the task of identifying AI-generated text is anticipated to become increasingly difficult in the future.}

\subsection{Attacks Against Forensics}
Several studies have demonstrated that detection systems are highly susceptible to paraphrasing-based attacks~\cite{sadasivan_can_2023}. Furthermore, recent developments reveal that more severe threats, such as LLMs, can be readily optimized to evade detection~\cite{kumarage2023reliable, nicks2023language}. These types of attacks present significant challenges to forensic analyses, necessitating more robust countermeasures in future iterations.

\subsubsection{LLM Variants}
\textls[-10]{The recent surge in open-source LLM development has unveiled a trend where the release of a powerful LLM is swiftly followed by numerous variations based on the same foundational model. These variants are produced through methods such as full fine-tuning, parameter-efficient tuning, or alignment approaches prevalent in the current LLM landscape. Often, these variations are specialized through training on domain-specific datasets or datasets generated by other advanced LLMs, like ChatGPT~\cite{gudibande2023false}. Noteworthy examples include the Alpaca~\cite{taori2023alpaca} and Vicuna~\cite{chiang2023vicuna} models, which are built on the Llama-base model. While these LLM variants contribute to the advancement of open-source LLM development, they present significant challenges to attribution and characterization systems when exploited by adversaries. For instance, these variants inherit the writing signatures of their base LLM, risking misattribution, potentially damaging the reputation of the original model's developers. Furthermore, malicious actors could craft their LLM variant by subtly incorporating harmful intent during the fine-tuning or alignment phases.}

\subsubsection{Coordinated AI Agents}
A significant trend within the current AI landscape involves the development of AI agents. These agents facilitate the deployment of powerful AI models that collaborate and operate autonomously to accomplish real-world tasks~\cite{park2023generative, murthy2023rex}. It is crucial to question whether existing frameworks are sufficiently equipped to detect, attribute, and characterize misinformation propagated by coordinated AI agents. In the future, we might encounter misinformation campaigns orchestrated by multiple LLMs working in concert. The effectiveness of existing forensic systems in addressing such threats remains an area that warrants further investigation.

\subsection{Towards Improved Forensic Systems}
\textls[0]{In today's AI era, the use of AI systems for text generation across diverse writing tasks is inevitable. Therefore, we anticipate a future where characterization emerges as the foremost element of AI-generated text forensics, i.e., the primary goal in safeguarding the information ecosystem will involve understanding malicious-intents behind AI-generations. Envisioning this future, we identify the following opportunities to enhance such forensic systems:}

\subsubsection{Knowledge-Aware LLMs} Advancing AI-generated text forensics could significantly benefit from integrating human expertise and existing forensic knowledge with LLM-based forensic systems~\cite{agrawal2023can}. By augmenting the LLMs using knowledge graphs~\cite{xu2022visualizing,zhang2023knowledge} that comprise human-expert forensic rules and knowledge, LLMs can used to build forensic systems that explain their decisions~\cite{chen2023models} accurately, which is crucial for characterization.

\subsubsection{Causality-aware Forensic Systems} 
\textls[-7]{From a characterization standpoint, forensic systems must extend beyond mere identification; it necessitates a deeper understanding of the underlying intent behind the generation such as the dissemination of false information or promotional material. To achieve this goal, we must address questions such as ``Why did the AI model generate this piece of text?'' and ``How would the text appear if it were generated with a different intent?''. Causality~\cite{pearl2009causality} answers ``why'' questions by explaining the relationships between events and allows us to examine alternative scenarios by considering different causal pathways and their potential consequences. Therefore, we believe Causality-aware AI-generated Text Forensic needs to be explored to thoroughly understand the underlying intent behind the text-generation and provide a holistic AI-generated text forensic system. 
This approach can be pursued in several directions, such as modeling the causal relationships between the AI model's training and input-output configurations, and causal reasoning to gain a deeper understanding of the text's intent.}

\section{Conclusion}

The field of AI-generated text forensics is rapidly evolving, with significant progress in detecting, attributing, and characterizing AI-generated texts. Current systems show promise in distinguishing between AI and human-written content, leveraging advanced techniques to analyze and identify subtle differences. However, the landscape is marked by ongoing challenges such as maintaining accuracy against the backdrop of rapidly improving AI technologies, ensuring adaptability to new types of generative models.

Looking forward, it's clear that the arms race between AI-generated text production and forensics will continue. The future of AI-generated text forensic research lies in enhancing the precision of existing tools, developing more dynamic models capable of adapting to new AI-generated text styles, and establishing ethical guidelines to govern the use and implications of these technologies. Ensuring the effectiveness of AI-generated text forensic systems against evolving AI capabilities will require a concerted effort from researchers, practitioners, and policymakers alike. 



\bibliography{paper}
\bibliographystyle{acl_natbib}

\onecolumn
\newpage
\appendix

\section{Additional Details: Experiment Settings and Benchmarks}
\label{sec:appendix}


\begin{table*}[h]
\resizebox{\textwidth}{!}{%
\begin{tabular}{p{4cm}p{0.8cm}p{3.5cm}p{6.65cm}p{9.5cm}p{1.8cm}p{2.9cm}}
\toprule
\rowcolor[HTML]{CBCEFB} 
\multicolumn{1}{c}{\cellcolor[HTML]{CBCEFB}} &
\multicolumn{6}{c}{\cellcolor[HTML]{CBCEFB}\textbf{Comparison Attributes}} \\ 
\cline{2-7}
\rowcolor[HTML]{CBCEFB} 
\multicolumn{1}{c}{\cellcolor[HTML]{CBCEFB}\textbf{Model}} & 
\multicolumn{1}{c}{\cellcolor[HTML]{CBCEFB}\textbf{Year}} &
\multicolumn{1}{c}{\cellcolor[HTML]{CBCEFB}\textbf{Generators}} &
\multicolumn{1}{c}{\cellcolor[HTML]{CBCEFB}\textbf{Domain}} &
\multicolumn{1}{c}{\cellcolor[HTML]{CBCEFB}\textbf{Data}} &
\multicolumn{1}{c}{\cellcolor[HTML]{CBCEFB}\textbf{Metrics}} &
\multicolumn{1}{c}{\cellcolor[HTML]{CBCEFB}\textbf{Top-Performance}} \\
\midrule
\rowcolor[HTML]{EFEFEF}
GLTR~\cite{gehrmann_gltr_2019} & 2019 & GPT-2 & 
\begin{tabular}[c]{@{}l@{}}News, Scientific Articles,\\ Childrens Books\end{tabular} & 
\begin{tabular}[c]{@{}l@{}}- Random paragraphs from the bAbI task children book corpus,\\ - New York Times articles (NYT),\\ - Scientific abstracts from nature and science (SA)\end{tabular} & 
AUROC & 0.87 \\
Fast-DetectGPT~\cite{bao_fast-detectgpt_2023} & 2023 & 
\begin{tabular}[c]{@{}l@{}}GPT-2, Neo-2.7 \end{tabular} & 
\begin{tabular}[c]{@{}l@{}}News, Wikipedia, Story Writing,\\ Translation, Medical QA\end{tabular} & 
\begin{tabular}[c]{@{}l@{}}- XSum, SQuAD, WritingPrompts,\\ - WMT'16, PubMedQA\end{tabular} & 
AUROC & 
\begin{tabular}[c]{@{}l@{}}0.9967, 0.9984 \end{tabular} \\
\rowcolor[HTML]{EFEFEF} 
AuthentiGPT~\cite{guo_authentigpt_2023} & 2023 & 
\begin{tabular}[c]{@{}l@{}}GPT-3.5, GPT-4\end{tabular} & 
Medical QA & 
\begin{tabular}[c]{@{}l@{}}- Human Generated articles from PubMedQA,\\ - Machine generated articles from PubMedQA\end{tabular} & 
\begin{tabular}[c]{@{}l@{}}Accuracy,\\ AUROC\end{tabular} & 
\begin{tabular}[c]{@{}l@{}}0.86,\\ 0.918\end{tabular} \\
OUTFOX~\cite{koike_outfox_2023} & 2023 & 
\begin{tabular}[c]{@{}l@{}}ChatGPT, GPT-3.5,\\ FLAN-T5-XXL\end{tabular} & 
Essays & 
- Machine generated essays & 
F1 & 
\begin{tabular}[c]{@{}l@{}}96.4,\\ 96.9,\\ 83.3\end{tabular} \\
\rowcolor[HTML]{EFEFEF} 
DetectGPT~\cite{mitchell_detectgpt_2023} & 2023 & T5-3B & 
\begin{tabular}[c]{@{}l@{}}News, Wikipedia, Story Writing,\\ Translation, Medical QA\end{tabular} & 
\begin{tabular}[c]{@{}l@{}}- XSum, SQuAD, WritingPrompts,\\ - WMT'16, PubMedQA\end{tabular} & 
AUROC & 97 \\
DetectLLM-LRR~\cite{su2023detectllm} & 2023 & T5-3B & 
\begin{tabular}[c]{@{}l@{}}News, Wikipedia, Story Writing\end{tabular} & 
\begin{tabular}[c]{@{}l@{}}- XSum, SQuAD, WritingPrompts\end{tabular} & 
AUROC & 92.7 \\
\rowcolor[HTML]{EFEFEF} 
GPT-who~\cite{venkatraman_gpt-who_2023} & 2023 & 
\begin{tabular}[c]{@{}l@{}}GPT-1, FAIR\_wmt20\end{tabular} & 
\begin{tabular}[c]{@{}l@{}}Author Attribution, Academic Articles,\\ Essays, Story generation\end{tabular} & 
\begin{tabular}[c]{@{}l@{}}- TuringBench Benchmark,\\ - GPA Benchmark,\\ - ArguGPT,\\ - DeepFake Text\end{tabular} & 
F1 & 
\begin{tabular}[c]{@{}l@{}}0.99, 0.99\end{tabular} \\
SeqXGPT~\cite{wang_seqxgpt_2023} & 2023 & 
\begin{tabular}[c]{@{}l@{}}GPT-2 XL, GPT-Neo,\end{tabular} & 
\begin{tabular}[c]{@{}l@{}}News, Social Media Posts, Medical QA,\\ Scientific Articles, Technical Documentation\end{tabular} & 
\begin{tabular}[c]{@{}l@{}}- XSum, IMDB, PubMedQA, arXiv, SQuAD\end{tabular} & 
\begin{tabular}[c]{@{}l@{}}Precision,\\ Recall\end{tabular} & 
99.2, 97.9, 99.3, 98.2 \\
\rowcolor[HTML]{EFEFEF} 
DetectGPT-SC~\cite{wang_detectgpt-sc_2023} & 2023 & ChatGPT & 
News & 
\begin{tabular}[c]{@{}l@{}}- CYN,\\ - Human ChatGPT Comparison Corpus\end{tabular} & 
Accuracy & 91.1 \\
GPT-4~\cite{bhattacharjee_fighting_2023} & 2023 & 
\begin{tabular}[c]{@{}l@{}}TRANSF\_XL, XLM\end{tabular} & 
News & 
- TuringBench & 
Accuracy & 
100,100 \\
\rowcolor[HTML]{EFEFEF} 
ChatGPT~\cite{zhu_beat_2023} & 2023 & ChatGPT & 
\begin{tabular}[c]{@{}l@{}}News, QA\end{tabular} & 
\begin{tabular}[c]{@{}l@{}}- MultiNews, GovReport, BillSum,\\ Finance, Reddit, Medicine\end{tabular} & 
AUROC & 90.05 \\
\bottomrule
\end{tabular}%
}
\caption{Zero-Shot Detection Models.}
\label{tab:zeroshot_styled}
\end{table*}

\begin{table*}[]
\resizebox{\textwidth}{!}{
\begin{tabular}{p{4cm}p{3cm}p{3cm}p{4cm}p{5cm}p{2.5cm}p{3cm}}
\toprule
\rowcolor[HTML]{CBCEFB} 
\multicolumn{2}{c}{\cellcolor[HTML]{CBCEFB}} & 
\multicolumn{5}{c}{\cellcolor[HTML]{CBCEFB}\textbf{Comparison Attributes}} \\ 
\cline{3-7}  
\rowcolor[HTML]{CBCEFB} 
\textbf{Model} &
\textbf{Generators} &
\textbf{Detection Models} &
\textbf{Domain} &
\textbf{Data Sources} &
\textbf{Metrics} & 
\textbf{Top Performance} \\ 
\midrule
\rowcolor[HTML]{EFEFEF} 
Energy-based model (EBM) \cite{bakhtin2019real} & GPT-2 & Linear, BiLSTM, UniTransf & News, Books, Wiki & Books: The Toronto books corpus, CCNews: De-duplicated subset of the English portion of the CommonCrawl news dataset, The wikitext103 dataset & Acc & 91.7\% on Books, 88.4\% on CCNews, 76.4\% on Wiki \\
Grover Detect \cite{zellers2019defending} & Grover & Grover, GPT-2, BERT, FastText & News & HW - April 2019 RealNews & Acc & 91.6\% on Grover-Mega \\
\rowcolor[HTML]{EFEFEF}
BERT-Classifier \cite{ippolito2020automatic} & GPT-2 & GLTR, BERT-Large, Bag-of-Words & Web & WebText Data (250k) & Acc (Model + Human evaluators) & Model: 90.1\%, Best Human Acc: ~85\% \\
BERT-GPT Ensemble \cite{adelani2020generating} & GPT-2, LSTM & Grover, GLTR, OpenAI GPT-2 & Reviews & Amazon and Yelp Reviews & EER & 20.9\% on Amazon, 19.6\% on Yelp \\
\rowcolor[HTML]{EFEFEF}
FAST \cite{zhong2020neural} & Grover, GPT-2 & RoBERTa & News, Web & Realnews, Webtext (OpenAI, Hugging face) & Acc & 84.9\% on News, 93.5\% on Webtext \\
STADEE \cite{chen2023stadee} & ChatGPT & RoBERTa & News, Finance, Medicine, Psychology & HC3-Chinese (In-Domain), ChatGPT-CNews (OODD), CPM-CNews (in-the-wild) & F1-Score & 87.05\% (In-domain), 87.4\% (OOD), Outperforms baseline by 9.28\% \\
\rowcolor[HTML]{EFEFEF}
Prompt-based Classification \cite{gagiano2023prompt} & T5, GPT-X & Falcon-7B & Law, Medicine & HW, MG English text, ALTA 2023 shared task dataset & Acc & 99.1\% on test data \\
J-Guard \cite{kumarage2023j} & Grover, CTRL, GPT-2, ChatGPT (3.5) & RoBERTa, BERT, DeBERTa, DistilBERT & News & TuringBench, ChatGPT generated news dataset & AUROC & 98.6\% on Grover, 96.8\% on ChatGPT \\
\rowcolor[HTML]{EFEFEF}
Fine-tuning and Semantic \cite{gambini2023detecting} & Bloom (1b7, 3b, 7b1), babbage, curie, text-davinci-003 & BERTweet, TriFuseNet & Legal, Web, News & Wiki, Tweets, Reviews & F1-score & BERTweet: 0.616, TriFuseNet: 0.715 \\
Attention Maps Topology \cite{kushnareva2021artificial} & GPT-2, Grover & BERT, TF-IDF & Web, Product Reviews, News & WebText, Amazon Reviews, RealNews & Acc & 87.7\% on WebText, 61.1\% on Amazon Reviews, 63.6\% on RealNews \\
\rowcolor[HTML]{EFEFEF}
CheckGPT \cite{liu2023check} & ChatGPT & RoBERTa & Academia - Research paper abstracts CS, Physics, Humanities, Social Sciences & GPABenchmark & Classification Accuracy & 98\% \\
GHOSTBUSTER \cite{verma2023ghostbuster} & ChatGPT, Claude & DetectGPT, GPTZero, RoBERTa & Student Essays, Creative Writing, News & subreddit, Reuters, IvyPanda articles & F1-score & 99 \\
\rowcolor[HTML]{EFEFEF}
Ensemble of Transformers \cite{mikros2023ai} & GPT & Ensemble (BERT, RoBERTa, ELECTRA, XLNet) & English Language & AuTexTification English corpora & Acc & 95.55\% \\
Stacking the Odds \cite{nguyen2023stacking} & GPT-X, T5 & ALBERT, ELECTRA, RoBERTa, XLNet & Law & 2023 Shared Task & Acc & 95.55\% \\
\rowcolor[HTML]{EFEFEF}
MGT Family and Scale \cite{sarvazyan2023supervised} & GPT-3 (babbage, curie, and davinci), BLOOM (1b7, 3b, 7b) & DeBERTa (En), MarIA (Spanish), XLM-RoBERTa, BLOOM-560M & English and Spanish language & AuTexTification corpus & F1-score & En: 85.6\% (BLOOM), 89.94\% (GPT), Es: 70.58\% (BLOOM), 94.97\% (GPT) \\
ConDA \cite{bhattacharjee2023conda} & CTRL, F19, GPT (G2X, G3), Grover\_Mega, XLM & RoBERTa & News & TuringBench4, ChatGPT News & F1-Score & Avg performance gains of 31.7\% from baseline \\
\rowcolor[HTML]{EFEFEF}
Human and AI-Generated Texts \cite{schaaff2023classification} & ChatGPT & GPTZero, ZeroGPT & Biology, Chemistry, Geography, History, IT, Music, Politics, Religion, Sports, Visual Arts & Human-AI-Generated Text Corpus (Mindner) & F1-score & 99\% for Spanish, 98\% for English, 97\% for German, and 95\% for French \\
Coco \cite{liu2023coco} & GroverMega, GPT-2 XLM-1542M, GPT-3.5 & RoBERTa, Attention LSTM & News, Web & Grover dataset, GPT-2 Dataset, GPT-3.5 Dataset & Acc, F1 & outperforms baseline by 2\% \\
\rowcolor[HTML]{EFEFEF}
DEMASQ \cite{kumari2023demasq} & ChatGPT & CheckGPT & Medicine, Finance, Social Media, Politics & Medicine, Open QA, arXiv, Political, Finance, Wiki, Social Media Posts & True Pos (TPR), True Neg (TNR) & TPR-97.0, TNR-96.5 \\
MFD \cite{wu2023mfd} & ChatGPT & Log Likelihood, Log Rank, Entropy, GLTR, DetectGPT, DetectLLM-LRR, MFD & Finance, Medicine, Open QA, Social Media Posts & Human ChatGPT Comparison Corpus & F1 & 98.41\% \\
\rowcolor[HTML]{EFEFEF}
LLMs for LLM Generated Text Detectors \cite{aguilar2023gpt} & BLOOM (1b7, 3b, 7b1), GPT-3 (Babbage, Curie, DaVinci-003) & BERT, RoBERTa, XLM-RoBERTa, DeBERTA, GPT-2 & Legal, News, Reviews, Tweets, How-to & MultiEURLEX, XSUM, Amazon Reviews, TSATC, WikiLingua & F1 & 92 \\
\bottomrule
\end{tabular}%
}
\caption{Supervised Detection Models (HW: Human Written, MG: Machine Generated).}
\label{tab:supervised_appendix}
\end{table*}

\begin{table*}[t]
\resizebox{\textwidth}{!}{
\begin{tabular}{p{3cm}p{4cm}p{3cm}p{5cm}p{4.5cm}p{2cm}p{3cm}p{2cm}p{3cm}}
\toprule
\rowcolor[HTML]{CBCEFB} 
\multicolumn{2}{c}{\cellcolor[HTML]{CBCEFB}} &
\multicolumn{7}{c}{\cellcolor[HTML]{CBCEFB}\textbf{Comparison Attributes}} \\ \cline{3-9}  
\rowcolor[HTML]{CBCEFB} 
\textbf{Model} &
\textbf{Generators} &
\textbf{Domain} &
\textbf{Training Samples} &
\textbf{Data Sources} &
\textbf{Metrics} & 
\textbf{Top Performance} &
\textbf{Multilingual} & 
\textbf{Seed Prompt}\\ 
\midrule
\rowcolor[HTML]{EFEFEF} 
Facts from Fiction \cite{mosca2023distinguishing} & SCIgen, GPT-2, GPT-3, ChatGPT, Galactica & Scientific Papers & arXiv & 16K (R), 13K (F), 4K (Para) & Acc & 77\% (OOD), 100\% (In-Domain) & \checkmark & Title, Abstract, Introduction as concatenated text \\ 
AuTexTification \cite{sarvazyan2023overview} & BLOOM (1B7, 3B, 7B1), GPT (babbage, curie, text-davinci003) & Tweets, Reviews, News, Legal, and How-to Articles & MultiEURLEX, XSUM, Amazon Reviews, TSATC, WikiLingua, Es(XLM-Tweets, MLSUM, COAR, COAH, TSD) & 160k texts & Macro-F1 & 80.91(En), 70.77(Es) & \checkmark & Domain-specific human-authored texts\\
\rowcolor[HTML]{EFEFEF} 
SAID \cite{cui2023said} & AI Users & Social Media & \textbf{Z}hihu and \textbf{Q}uora & \textbf{Q}(HW-14648, MG-22892), \textbf{Z}(HW-72565, MG-108654) & Acc & 96.50\%  & \checkmark & Text modification, paraphrasing \\
GPABenchmark \cite{liu2023check} & ChatGPT & CSE Tech, Physics, Humanities, Social Sciences Writing & Research Paper Abstracts & 600K (HW + MG) & Acc & 98\% & $X$ & Review, Polish, Revise, Rewrite and Edit the Title, Abstract\\
\rowcolor[HTML]{EFEFEF} 
Academic Text \cite{liyanage2023detecting} & GPT & Academia & DAGPap22, GPT Wiki Intro & 500-R and 500-MG & F1-Score & 97.5\% & $X$ & first 7 words of Wiki Intro, first 50 Words of Academic paper or first sentence of  Abstract \\
MULTITuDE \cite{macko2023multitude} & Multilingual LLMs - GPT-3, GPT-4, LLaMA65B, ChatGPT, Alpaca-LoRa-30B, Vicuna-13B, OPT-66B, IML-Max-1.3B & News & MassiveSumm & 11 languages- 74K (8K-HW, 66K-MG)  & Acc & 94.50\% (En) & \checkmark & Titles of selected articles \\
\rowcolor[HTML]{EFEFEF} 
To ChatGPT \cite{pegoraro2023chatgpt} & ChatGPT & Medical, Open QA, Finance  & User-generated responses from popular Social Networking Platforms & 131K (58k HW, 72K MG) & TPR\% (Detection Capability) & Detects 90\% as HW & $X$ &  Inquiry prompts\\
H3Plus \cite{su2023hc3} & ChatGPT & News & CNN,DailyMail, Xsum, LCSTS, News2016, WMT & 210K (42K-Chinese, 95K-En Train samples) & Acc & 99.5\% En, 98.65\% Chinese & \checkmark & Translate, Summarize, and Paraphrase original text \\
\rowcolor[HTML]{EFEFEF} 
TURINGBENCH \cite{uchendu2021turingbench} & GPT-1, GPT-2, GPT-3, PPLM, Transformer\_XL, XLNET, Grover, CTRL, XLM, FAIR & Politics, News  & CNN, Washington Post & 10K R, 200K MG & F1-score & 87.9(Detection), 81(Attribution) & $X$ & Article Title \\
M4 \cite{wang2023m4} & ChatGPT, textdavinci-003, LLaMa, FlanT5, Cohere, Dolly-v2, BLOOMz & News, Scientific Article Peer Reviews, Social Media, Web, History, Science & Wiki, WikiHow, Reddit, arXiv (En), PeerRead, Baike, WebQA(Chinese), News(Urdu, Indonesian), RuATD (Russian)  & 122K(En-101K, other Lang-9K each) & F1-Score & 99.7\%  & \checkmark & News Title and Headlines, Paper Abstract and Title, Question Title and Description\\
\rowcolor[HTML]{EFEFEF} 
GHOSTBUSTER \cite{verma2023ghostbuster} & ChatGPT, Claude & Student Essays, Creative Writing, News & subreddit, Reuters, IvyPanda articles & 21K (1K HW, 6K MG (5K ChatGPT, 1K Claude)) per domain & F1-score & 99 & $X$ & Length, Headline and Document \\
HPPT \cite{yang2023chatgpt} & ChatGPT & Scientific Papers & HW abstracts of accepted papers from NLP academic conferences & 6050 Abstracts (R), & Acc & 94.5\% & $X$ & Abstracts \\
\midrule
\multicolumn{5}{c}\textbf{Misinformation} \\
\midrule
\rowcolor[HTML]{EFEFEF} 
LLMFake \cite{chen2023can} & ChatGPT, Llama2 (7b,13b,70b), Vicuna (7b,13b,33b) & News, Healthcare, Politics & \textbf{Pol}itifact, \textbf{Gos}sipcop, \textbf{CoA}ID  & \textbf{Pol}(270 NF, 145 F), \textbf{Gos}(2230 NF), \textbf{CoA}(925 F) & Success Rate & Drops by 19\% & $X$ & Collect 100 pieces of misinformation \\
F3 \cite{lucas2023fighting} & GPT-3.5 & Political, News, Social Media & Politifact1, Snopes & HW (5508R, 7215F), MG (9141R, 18526F) & Acc & 72\% & $X$ & Standard Impersonator, Dataset Content, Instruct to paraphrase, rephrase and reword the content\\
\rowcolor[HTML]{EFEFEF}
ODQA-NQ-1500, CovidNews \cite{pan2023risk}  & GPT-3.5 (text-davinci-003) & Web, News & Wiki Natural Questions, StreamingQA News & 21M (NQ), 3.3M (Cov) & Exact Match & 87\% Drop & $X$ & Generate content to answer questions like human-written factual article \\
Covid-19 Misinfo \cite{zhou2023synthetic} & GPT-3.5 & News, Social Media (SM) & COVID19-FNIR, COVID Rumor, Constraint & 12k (6768 News, 5640 SM) & F1-score & 98.5 & $X$ & COVID-19-related keywords - virus and outbreak \\
\rowcolor[HTML]{EFEFEF}
\textbf{Gos}sipCop++, \textbf{Pol}itiFact++ \cite{su2023fake} & ChatGPT & News, Politics & FakeNewsNet, PolitiFact, GossipCop & 10K HW (5K-R, 5K-F), 5K-MF & Acc & 88\% \textbf{Gos++}, 80.93\% \textbf{Pol++} & $X$ & Title and Description \\
\rowcolor[HTML]{EFEFEF}
D-Human \cite{jiang2023disinformation} & ChatGPT (3.5, 4) & News, Politics & Reuters, Politifact & 21K-R, 23K-F, 23K-MG & Misclassified \% & 77.93\% & $X$ & Summary with Role and Tone, Extract all keywords and assume the role of Journalist. Rewrite original text in 3 versions \\
\rowcolor[HTML]{EFEFEF}
 \textbf{HANSEN}~\cite{tripto2023hansen} &  ChatGPT, PaLM2, Vicuna13B & human spoken conversations - Youtube, Movie-Dialogs & HANSEN (from 17 human datasets - TED, SEC, Spotify, CEO, Tennis etc.) & 21k & Authorship attribution ( macroF1), Verification (Auc) & 0.98  & \checkmark & Speech Transcripts, Talk show Titles, conversation utterances.  \\
\bottomrule
\end{tabular}%
}
\caption{Benchmark Datasets (R: Real, F: Fake, Para: Paraphrased, F:  Factual, NF:  NonFactual, HW:  Human Written, MG:  Machine Generated, HR: Human Real, HF: Human Fake, MF: Machine Fake, En: English, Es: Spanish).}
\label{tab:benchmark_appendix}
\end{table*}

\end{document}